\documentclass[conference]{IEEEtran}
\usepackage{times}

\usepackage[numbers,sort&compress]{natbib}
\usepackage{multicol}

\usepackage[dvipsnames]{xcolor}

\usepackage{amsthm}
\usepackage{amsmath}
\usepackage[font=footnotesize,labelfont=bf]{caption}
\usepackage[font=footnotesize,labelfont=bf]{subcaption}
\usepackage{float}
\usepackage{algorithm}
\usepackage{algpseudocode}
\usepackage{duckuments}
\usepackage{graphicx}
\usepackage{xr}
\let\labelindent\relax
\usepackage{enumitem}
\usepackage{etoolbox}
\usepackage{mathrsfs}
\usepackage{multirow}
\usepackage{overpic}
\usepackage{nccmath}
\usepackage{titlesec}
\usepackage[nomath]{cellspace}
\usepackage{soul} 

\DeclareCaptionType{CapEq}[][]
\definecolor{tab:blue}{RGB}{31,119,180}
\definecolor{tab:orange}{RGB}{255,127,14}
\definecolor{tab:green}{RGB}{44,160,44}
\definecolor{tab:red}{RGB}{214,39,40}
\definecolor{tab:purple}{RGB}{148,103,189}

\definecolor{sns:blue}{rgb}{0.00392156862745098, 0.45098039215686275, 0.6980392156862745}
\definecolor{sns:orange}{rgb}{0.8705882352941177, 0.5607843137254902, 0.0196078431372549}
\definecolor{sns:green}{rgb}{0.00784313725490196, 0.6196078431372549, 0.45098039215686275}
\definecolor{sns:red}{rgb}{0.8352941176470589, 0.3686274509803922, 0.0}
\definecolor{sns:purple}{rgb}{0.8, 0.47058823529411764, 0.7372549019607844}
\definecolor{sns:brown}{rgb}{0.792156862745098, 0.5686274509803921, 0.3803921568627451}
\definecolor{sns:pink}{rgb}{0.984313725490196, 0.6862745098039216, 0.8941176470588236}
\definecolor{sns:gray}{rgb}{0.5803921568627451, 0.5803921568627451, 0.5803921568627451}
\definecolor{sns:yellow}{rgb}{0.9254901960784314, 0.8823529411764706, 0.2}
\definecolor{sns:light_blue}{rgb}{0.33725490196078434, 0.7058823529411765, 0.9137254901960784}
\newtheoremstyle{dense}
  {3pt} 
  {3pt} 
  {\itshape} 
  {} 
  {\bfseries} 
  {:} 
  {.5em} 
  {} 
\theoremstyle{dense}

\newtheorem{remark}{Remark}
\AtBeginEnvironment{remark}{\small}
\AtBeginEnvironment{quote}{\par\singlespacing\small}
\usepackage{stix}
\usepackage{tikz}
\DeclareRobustCommand{\thickX}{
    \begin{tikzpicture}[baseline=0ex, line width=2, scale=0.13];
    \draw (0,0) -- (1,1);
    \draw (0,1) -- (1,0);
    \end{tikzpicture}}

\newcommand{\SymiMESA}{\textcolor{sns:blue}{$\bigstar$}}
\newcommand{\SymCentralized}{\textcolor{sns:gray}{$\mdblkcircle$}}
\newcommand{\SymDLGBP}{\textcolor{sns:orange}{$\mdblksquare$}}
\newcommand{\SymDLGBPWindowed}{\textcolor{sns:red}{$\blacktriangle$}}
\newcommand{\SymDDFSAM}{\textcolor{sns:green}{$\thickX$}}
\newcommand{\SymIndependent}{\textcolor{sns:pink}{$\blacklozenge$}}
\usepackage[bookmarks=true]{hyperref}
\usepackage{cleveref}
\hypersetup{
    colorlinks=true,
    linkcolor=blue,
    filecolor=magenta,      
    urlcolor=cyan,
    pdfpagemode=FullScreen,
}
\AtBeginEnvironment{tabular}{\footnotesize}
\titleformat{\subsubsection}[runin]
   {\itshape}
   {\thesubsubsectiondis}
   {0.5em}
   {}
   [:]
\titlespacing\subsubsection{\parindent}{0pt}{0.5em}
\titlespacing\subsection{0pt}{6pt plus 2pt minus 2pt}{2pt plus 2pt minus 2pt}

\usepackage{amsfonts}
\usepackage{amsthm}

\DeclareMathOperator*{\argmax}{arg\,max}
\DeclareMathOperator*{\argmin}{arg\,min}

\newcommand{\logmap}[1]{\mathrm{Log}\left(#1\right)}

\newcommand{\norm}[1]{\left\|#1\right\|}

\newcommand{\Cc}{\mathcal{C}}

\newcommand{\Ec}{\mathcal{E}}
\newcommand{\Fc}{\mathcal{F}}
\newcommand{\Gc}{\mathcal{G}}

\newcommand{\Kc}{\mathcal{K}}

\newcommand{\Mc}{\mathcal{M}}

\newcommand{\Rc}{\mathcal{R}}
\newcommand{\Sc}{\mathcal{S}}

\newcommand{\Zc}{\mathcal{Z}}

\begin{document}

\title{\huge iMESA: Incremental Distributed Optimization for Collaborative Simultaneous Localization and Mapping \vspace{-4mm}}


\author{
\authorblockN{Daniel McGann and Michael Kaess}
\authorblockA{
The Robotics Institute, 
Carnegie Mellon University, Pittsburgh PA, USA\\}
Email: \texttt{\{danmcgann, kaess\}@cmu.edu}
\vspace{-3mm}
}
\twocolumn[{
	\renewcommand\twocolumn[1][]{#1}
	\maketitle
	\thispagestyle{empty}
	\begin{center}
		\vspace{-6mm}
		\centering
        \includegraphics[width=\linewidth]{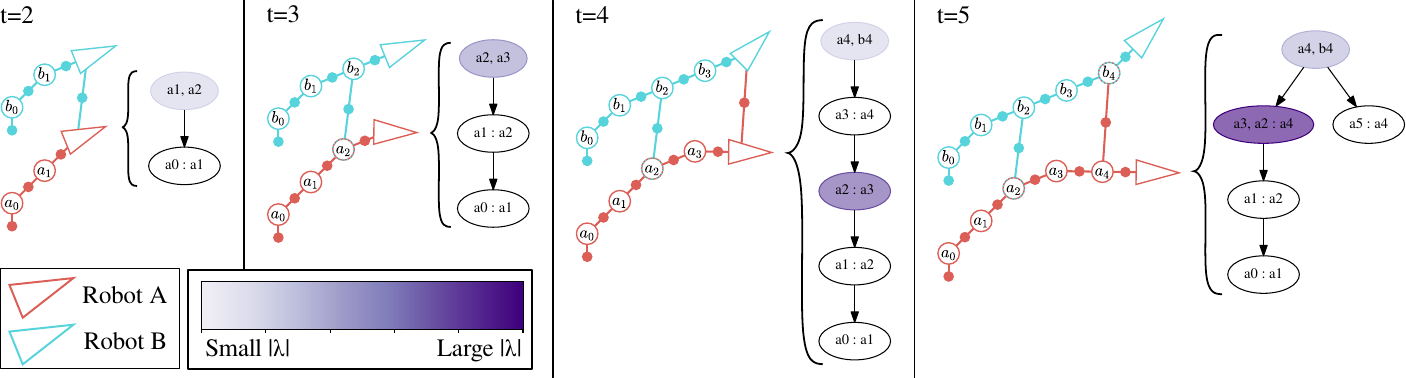}
        \captionof{figure}{
        An illustration of the iMESA algorithm. When robots observe shared variables, they enforce equality using "biased priors" in their local factor-graph. Biased priors control constraint tightness with dual variables ($\lambda$) which are shown for robot \textcolor[HTML]{db5f57}{A} as \textcolor[HTML]{54278f}{purple} in the robot's Bayes Tree. Over time, as communication is available, robots tighten equality constraints with dual gradient ascent to provide consistent solutions. Meanwhile, robots incorporate new measurements efficiently using the iSAM2 algorithm. Through this process, iMESA is able to accurately and efficiently solve incremental distributed C-SLAM problems.
        }
        \label{fig:amortization_explination}
		\vspace{-1mm}
	\end{center}
}]

\begin{abstract}
This paper introduces a novel incremental distributed back-end algorithm for Collaborative Simultaneous Localization and Mapping (C-SLAM). For real-world deployments, robotic teams require algorithms to compute a consistent state estimate accurately, within online runtime constraints, and with potentially limited communication. Existing centralized, decentralized, and distributed approaches to solving C-SLAM problems struggle to achieve all of these goals. To address this capability gap, we present \underline{I}ncremental \underline{M}anifold \underline{E}dge-based \underline{S}eparable \underline{A}DMM (iMESA) a fully distributed C-SLAM back-end algorithm that can provide a multi-robot team with accurate state estimates in real-time with only sparse pair-wise communication between robots. Extensive evaluation on real and synthetic data demonstrates that iMESA is able to outperform comparable state-of-the-art C-SLAM back-ends.  

\end{abstract}
\IEEEpeerreviewmaketitle


\section{Introduction}\label{sec:intro}

Collaborative Simultaneous Localization and Mapping (C-SLAM) is a fundamental capability for multi-robot teams~\cite{lajoie_cslam_survey_2022}. A key component of the C-SLAM system is the back-end algorithm responsible for estimating the state of the robot team from their distributed, noisy measurements~\cite{slam_survey_leonard_2016}. However, existing C-SLAM back-end algorithms struggle to handle the practical conditions experienced by multi-robot teams deployed in the real-world. During field deployments (e.g. search and rescue, forestry inspection, or scientific exploration~\cite{multirobot_sar_drew_2022, yuan_fire_survey_2015, cadre_mission_website}), multi-robot teams cannot assume the presence of communication infrastructure. Rather, teams can only assume an ad-hoc network that permits sparse communication between robots. Additionally, we expect multi-robot teams will need to re-plan frequently as new information is gathered. Such re-planning requires that the team have accurate and up-to-date state estimates. Existing C-SLAM back-end algorithms struggle to operate under these conditions either requiring reliable network connectivity, needing impractical amounts of computation time, or struggling to achieve accurate results. 

In this paper, we work to bridge the gap between existing C-SLAM back-end capabilities and the requirements for real-world multi-robot teams. We do so by developing an incremental distributed C-SLAM back-end that can tolerate sparse communication between robots and can provide accurate state estimates to the robot team in real-time.

The rest of the paper is structured as follows. First, we concretely define the incremental, communication limited, and computationally constrained C-SLAM problem we strive to address. Next, we review prior works and discuss their applicability to solve this task. We then provide a summary of the prior work MESA~\cite{mesa_mcgann_2023} before introducing our novel incrementalization of that algorithm -- iMESA (Fig.~\ref{fig:amortization_explination}). We conclude by rigorously evaluating the proposed algorithm on synthetic and real-world data to demonstrate that it achieves high quality results and is able to outperform existing algorithms. Finally, to aid future research, we make the source code for iMESA publicly available online\footnote{Source Code: \url{https://github.com/rpl-cmu/imesa}}.

\section{Problem Definition}\label{sec:problem_definition}
We seek to solve the generic C-SLAM problem in which a team of robots $\Rc$ estimates variables of interest $\Theta$ using noisy measurements $\Mc$. In this multi-robot case, each robot takes a subset of the measurements and observes a subset of the total variables. Importantly, some measurements taken by robots will be inter-robot measurements. These measurements may come from direct observation of other robots or from a distributed place recognition system~\cite{distrbuted_loopclosure_tian_2020}. Inter-robot measurements enable the multi-robot team to collaborate. This can improve their individual state estimates and can ensure all robots' solutions remain in the same global frame. Due to inter-robot measurements, multiple robots may observe the same variable. Therefore, let $\Theta_i \subset \Theta$ denote the non-disjoint subset of variables observed by robot $i\in \Rc$. Additionally, let $\Mc_i \subset \Mc$ denote the robot's disjoint subset of measurements. 

The de-facto standard for SLAM is to formulate the problem as a factor-graph and solve for the optimal variables via Maximum A-Posteriori (MAP) inference~\cite{slam_survey_leonard_2016}:

\begin{equation} \label{eq:cslam_as_map}
    \Theta_{MAP} = \argmax_{\Theta \in \Omega} P(\Theta) \prod_{i\in \Rc} P(\Mc_i | \Theta_i)
\end{equation}

where $\Omega$ is the product manifold constructed by the manifold of each variable in $\Theta$. When we assume that each measurement $m\in \Mc$ is affected by Gaussian noise with covariance $\Sigma_m$ this problem can be solved by Nonlinear Least Squares (NLS) optimization~\cite{factor_graphs_for_robot_perception}:

\begin{equation} \label{eq:cslam_as_nls}
    \Theta_{MAP} = \argmin_{\Theta \in \Omega} \sum_{i\in\Rc}\sum_{m\in \Mc_i} \norm{h(\Theta_i) - m}_{\Sigma_m}^2
\end{equation}

where $h(\Theta_i)$ is the measurement prediction function that computes the expected measurement from an estimate of the state. However, we are not interested in just solving this optimization problem once. For practical robotic applications we need to solve this problem at every timestep, with the advantage that we have access to the previous solution. We refer to this problem as the incremental C-SLAM problem to distinguish it from the "batch" problem above~\cite{isam_kaess_2007}:

\begin{equation}
\label{eq:incremental_cslam_as_nls}
\begin{aligned}
\Theta^t_{MAP} =& \argmin_{\Theta^t \in \Omega^t} \sum_{i\in\Rc}\sum_{m\in \Mc_i^t} \norm{h(\Theta^t_i) - m}_{\Sigma_m}^2 \\
\textrm{given:} \quad & \Theta_{MAP}^{t-1} \\
\end{aligned}
\end{equation}

where \eqref{eq:incremental_cslam_as_nls} must be solved online to report results with minimal delay to downstream tasks like planning.

\begin{remark}[Generic C-SLAM vs. PGO] \label{remark:generic_cslam_vs_pgo}
It is common in state estimation literature for C-SLAM back-ends to solve only Pose-Graph Optimization (PGO). PGO is a subset of C-SLAM in which each variable is a pose that lives on the SE$(N)$ manifold and all measurements are relative poses~\cite{dgs_choudhary_2017, geod_cristofalo_2020, asapp_tian_2020, dc2pgo_tian_2021, maj_min_iros_fan_2020, maj_min_journal_fan_2024}. PGO is useful in many applications and provides structure to exploit in algorithm design. However, it limits map representations (e.g. landmarks) as well as measurement sources (e.g. bearing and range), the latter of which is particularly useful in specific domains (e.g. underwater~\cite{auv_nav_review_paull_2014} and space~\cite{boroson_puffer_range_pgo_2020}). To develop the most useful C-SLAM back-end across all applications, we focus on the generic C-SLAM problem.
\end{remark}

In the incremental C-SLAM setting, information is distributed across robots in the team and robots must communicate to solve \eqref{eq:incremental_cslam_as_nls}. In real-world applications, the team may not have access to existing communication infrastructure. As such, we can only reliably assume that robots can communicate over an ad-hoc network built from the robots themselves. Due to the scale of the environment, hardware constraints, motion of the robots, and sources of interference, this network will be bandwidth limited, time varying, and frequently disconnected. 

Let us represent the communication network at a specific time $t$ by a undirected graph $\Gc^t = (\Rc, \Ec^t)$ with nodes made up of the robots $\Rc$ and edges $\Ec^t$ defined by the currently available connections between robots. Moreover, we assume that at each timestep any robots connected in $\Gc^t$ are only able to communicate pairwise with one other robot during that timestep due to bandwidth and computation constraints.
 
This communication model is very restrictive and represents a worst-case scenario. However, it is representative of what multi-robot teams may encounter in remote wilderness, disaster areas, or on extraterrestrial bodies. Additionally, an algorithm that can operate effectively under this restrictive model can also operate in more optimistic scenarios. If the team has use of communication infrastructure, then $\Gc^t$ is simply more densely connected at each timestep. If network bandwidth and robot hardware permits communication with multiple teammates, the algorithm can be run multiple times per timestep to use the additional resources and converge faster. Therefore, algorithms that can tolerate this model will be applicable even when communication is less limited.
\section{Related Work}\label{sec:related-work}

Existing works struggle to provide effective solutions to incremental, communication-limited C-SLAM problems.

The simplest approach to the task is to use a centralized back-end. In this method, robots communicate their measurements to a central server that performs all computation and sends solutions back to the team. These approaches can make use of existing incremental solvers like iSAM2 or MR-iSAM2 to provide very accurate results~\cite{isam2_kaess_2012, mrisam2_zhang_2021}, and have been used successfully under a variety of conditions~\cite{decent_loc_bailey_2011, mr3dlidar_slam_dube_2017, multi_uav_slam_schmuck_2017, cvi_slam_karrer_2018, ccm_slam_schmuck_2019, covins_schmuck_2021, dist_client_server_slam_zhang_2021, lamp2_chang_2022, cvids_zhang_2022, hydramulti_chang_2023, maplab_cramariuc_2023}. However, they require that $\Gc^t$ is connected at each timestep and can support significant network traffic to return solutions to all team members. Additionally, a central server introduces a single point of failure into the system and centralization does not scale well as the number of robots increases~\cite{subt_lessons_2023}.

Similar to centralized algorithms are decentralized back-ends. In these approaches each robot holds a copy of the global problem and independently computes the solution~\cite{decent_auv_paull_2014, mr6d_slam_schuster_2015, mr6d_slam_journal_schuster_2019, dense_mr_slam_dubois_2020, disco_slam_haung_2022, swarm_slam_lajoie_2024}. Like centralized methods, decentralized back-ends can use existing solvers to provide accurate results. Furthermore, decentralization removes any single point of failure from the system. However, decentralized back-ends require even greater bandwidth than centralized methods to send all measurements to every other robot. Additionally, decentralization necessitates that each robot performs expensive and redundant computation. These issues with bandwidth and computation can be mitigated, but not removed, by sparsifying the global problem~\cite{mr6d_slam_schuster_2015} and sharing computation within robot clusters~\cite{swarm_slam_lajoie_2024}.

Another major class of C-SLAM back-ends are distributed algorithms. Choudhary et al. proposed distributed Multi-Block ADMM (MB-ADMM) for this task~\cite{choudhary_admm_2015}. MB-ADMM can solve general C-SLAM problems accurately under some conditions. However, it is not guaranteed to converge for even linear problems and it requires synchronous communication over a connected network at each timestep~\cite{mbadmm_bad_chen_2016}.

Choudhary et al. also proposed the Distributed Gauss Sidel (DGS) algorithm for solving distributed batch PGO~\cite{dgs_choudhary_2017}. DGS optimizes a chordal relaxation of the PGO problem to provide robust but approximate solutions~\cite{chordal_init_carlone_2015}. Despite recovering only approximate solutions and requiring reliable communication, DGS has seen success in real-world systems~\cite{efficient_dist_vis_slam_cieslewski_2018, doorslam_lajoie_2020, dcl_slam_zhong_2023}.

To avoid the approximations made by DGS, Cristofalo et al. proposed the GeoD algorithm which optimizes the "Geodesic" PGO formulation enabling better accuracy~\cite{geod_cristofalo_2020}. GeoD treats each pose as an independent agent and optimizes the objective using gradient descent. GeoD has been shown to provide more accurate results than DGS at the cost of slower convergence.

Also targeting distributed batch PGO, Fan and Murphy proposed a Majorization-Minimization algorithm (MM-PGO) to optimize geodesic PGO~\cite{maj_min_iros_fan_2020, maj_min_journal_fan_2024}. This approach performs optimization via gradient descent and can guarantee convergence to first order critical points. While this optimization has slow convergence it can be sped up using Nesterov's method.

Going one step further than guaranteeing convergence to first order critical points, Tian et al. proposed an algorithm that optimizes the semi-definite relaxation of the batch PGO problem using distributed Riemannian gradient descent~\cite{dc2pgo_tian_2021, asapp_tian_2020}. The algorithm (DC2-PGO) and its asynchronous variant (ASAPP) can both compute accurate solutions to PGO and can even certify correctness of the solution under some conditions~\cite{sdps_holmes_2023}. These algorithms have been used with success on real world systems~\cite{kimeramulti_yun_2021, kimeramulti_journal_tian_2022, kimeramulti_lessons_tian_2023}. However, they demonstrate slow convergence due to their use of first order optimization.

Recently, McGann et al. proposed MESA, a distributed optimization algorithm for generic batch C-SLAM problems based on Consensus Alternating Direction Method of Multipliers (C-ADMM)~\cite{mesa_mcgann_2023}. In addition to handling generic C-SLAM problems, MESA demonstrates significantly faster convergence and improved or comparable accuracy to other works.

However, MB-ADMM, GeoD, MM-PGO, DC2-PGO, ASAPP, and MESA are all insufficient to solve incremental, communication-limited C-SLAM problems. All of these distributed algorithms target batch C-SLAM problems. Naively, these methods could be applied to the incremental case \eqref{eq:incremental_cslam_as_nls} by solving each timestep as a batch problem. However, doing so cannot meet the requirements outlined in Sec.~\ref{sec:problem_definition}. These methods require hundreds to thousands of iterations to converge, significantly exceeding the communication and computation requirements of our problem. Further, while some of these methods can tolerate asynchronous communication and thus communication dropout between algorithm iterations, they would all require a connected network during the current timestep, which is not guaranteed in real world scenarios.

Due to the challenges faced by distributed batch back-ends, some prior works have proposed distributed algorithms specifically targeting incremental C-SLAM. Cunningham et al. proposed Distributed Data Fusion Smoothing and Mapping (DDF-SAM and DDF-SAM2) which are custom distributed optimization algorithms designed around sharing marginals without double counting information~\cite{ddfsam_cunningham_2010, ddfsam2_cunningham_2013}. DDF-SAM2 is able to achieve online performance. However, it does not constrain robots to maintain equal linearization points which can significantly degrade solution consistency and accuracy. 

Recently, Murai et al. proposed a new incremental distributed C-SLAM back-end based on Distributed Loopy Gaussian Belief Propagation (DLGBP)~\cite{robot_web_journal_murai_2024}. DLGBP has been shown to recover very accurate results under many problem conditions, but struggles to converge in some settings. Additionally, DLGBP achieves online performance with a windowed approach in which only the $k$ most recent factors are updated. This windowing can significantly degrade performance, particularly when robots observe loop-closures.

Concurrent to this work, Matsuka and Chung also proposed a distributed incremental factor-graph optimizer based on C-ADMM called iDFGO~\cite{idfgo_matsuka_2023}. While iDFGO was designed for optimizing convex multi-agent factor-graphs with linear constraints, it can be applied to nonlinear problems like C-SLAM. While similar in spirit to the algorithm we propose, there are a number properties of this concurrent work (e.g. required synchronous communication) that make it difficult to apply to incremental, communication limited C-SLAM problems. For details on these properties see Remark~\ref{remark:idfgo_imesa_differences}.

Within the context of the problem defined in Sec.~\ref{sec:problem_definition} we can see that existing works do not provide a solution that is sufficient. Centralized and decentralized approaches cannot tolerate the desired communication model and scale poorly with the number of robots in the team. Similarly, batch distributed algorithms like MB-ADMM, DGS, GeoD, MM-PGO, DC2-PGO, ASAPP, and MESA cannot meet either the communication or computational requirements set forth. Even iDFGO, which shares a theoretical basis with our algorithm, cannot operate within bounds of our problem definition. While DDF-SAM2 and DLGBP are able to function within the bounds of our problem, their theoretical drawbacks often result in inaccurate solutions. To fill the gap left by prior works, we propose a novel incremental C-SLAM back-end -- iMESA.

\section{Manifold Edge-based Separable ADMM (MESA)}\label{sec:mesa}
In this section we review the batch distributed C-SLAM back-end algorithm MESA as we build off this prior work to design our proposed algorithm iMESA. For additional details on MESA, refer to the original publication~\cite{mesa_mcgann_2023}. MESA models batch C-SLAM \eqref{eq:cslam_as_nls} as a constrained optimization problem:

\begin{equation} \label{eq:mesa_cslam_as_cosntrained_opt}
\begin{aligned}
    \Theta_{MAP} = \argmin_{\substack{\Theta_i \in \Omega_i \forall i\in\Rc\\ Z \in \Zc}}\quad& 
        \sum_{i\in \Rc} \sum_{m\in \Mc_i} \norm{h_m(\Theta_i) - m}_{\Sigma_m}^2 \\
        \textrm{s.t.} \quad & \logmap{\theta_{s_i} \ominus z_{(i,j)_s}} = 0 \\
        & \logmap{\theta_{s_j} \ominus z_{(j,i)_s}} = 0 \\
        & ~\forall~ \theta_s \in \Sc_{(i,j)} ~\forall~(i,j) \in \Rc\times\Rc\\
\end{aligned}
\end{equation}

where $\Omega_i$ is the product manifold to which $\Theta_i$ belongs, $S_{(i,j)}$ is the set of variables (e.g. robot poses or landmarks in the environment) shared between robots $i$ and $j$, $\theta_{s_i}$ is the copy of shared variable $s$ owned by robot $i$. For each shared variable, the problem is augmented with edge variables $z_{(i,j)_s}$ and $z_{(j,i)_s}$ which are held by robots $i$ and $j$ respectively. $Z$ is the set of all $z_{(i,j)_s}$ and $\Zc$ is the appropriate product manifold that is further constrained such that $z_{(i,j)_s} = z_{(j,i)_s}$.

Constraints are introduced, as inter-robot measurements result in some state variables that are shared between robots. To ensure the multi-robot teams produces a consistent solution, we must constrain the local solutions to these variables to be equal between all robots that share them. Additionally, variables in C-SLAM problems are typically robot poses on the Special Euclidean manifold. Thus, constraints may be nonlinear and are written using Lie group notation. A tutorial on Lie groups and this notation is provided by Solà et al.~\cite{microlie_sola_2021}.

\begin{remark}[Constraint Functions] \label{remark:constraint_functions}
The authors of MESA originally proposed a general formulation of \eqref{eq:mesa_cslam_as_cosntrained_opt} using a generic constraint function. Given existing results, we opt to use Geodesic constraints~\cite[Table~1]{mesa_mcgann_2023}. However, any valid constraint function could be used.
\end{remark}

MESA solves this constrained optimization problem via separable C-ADMM using an edge-based communication model and taking special considerations for on manifold variables. Concretely, this approach solves the Augmented Lagrangian for problem \eqref{eq:mesa_cslam_as_cosntrained_opt} by alternating between optimizing the primal variables $\Theta$ and performing gradient ascent on the dual variables $\Lambda$. The augmentation with edge variables $z_{(i,j)_s}$ permits the C-ADMM iterates to be solved in an asynchronous distributed fashion. The iterates are as follows:

\begin{align}
    & \label{eq:mesa_updates_x}
    \begin{aligned}
    \Theta^{k+1}_{i}& = \argmin_{\Theta_i \in \Omega_i} \quad 
          \sum_{m\in \Mc_i} \norm{h_m \left(\Theta^k_i\right) - m}_{\Sigma_m}^2 \\
          &+ \sum_{j \in \Rc} \sum_{s\in \Sc_{(i, j)}}
             \frac{\beta^k_{(i,j)}}{2} \norm{\logmap{\theta^k_{s_i} \ominus z^k_{(i,j)_s}}  + \frac{\lambda^k_{(i,j)_s}}{\beta^k_{(i,j)}}}^2 
    \end{aligned}\\
    & \label{eq:mesa_updates_z}
    \begin{aligned}
         {z^{k+1}_{(i, j)_s}}& = \mathrm{SPLIT}\left(\theta^{k+1}_{s_i}, \theta^{k+1}_{s_j}, 0.5\right)
    \end{aligned}\\
    & \label{eq:mesa_updates_dual}
    \lambda_{(i,j)_s}^{k+1} = \lambda_{(i,j)_s}^{k} + \beta_{(i,j)}^k \logmap{\theta^{k+1}_{s_i} \ominus z^{k+1}_{(i,j)_s}} \\
    \label{eq:mesa_updates_beta}
    & \beta_{(i,j)}^{k+1} = \alpha \beta_{(i,j)}^k
\end{align}

where $k$ is the iteration number, $\lambda_{(i,j)_s} \in \Lambda$ is the dual variable added for the constraint between $\theta_{s_i}$ and $ z_{(i,j)_s}$, $\beta_{(i,j)}$ is the penalty parameter and learning rate for gradient ascent of the dual variables, $\alpha$ is a parameter to scale this penalty parameter, and $\mathrm{SPLIT}(a, b, \gamma)$ interpolates $a$ and $b$ by factor $\gamma \in [0,1]$ such that any linear component is interpolated linearly, and any rotational component is interpolated spherically.

The bulk of computation in MESA occurs in iterate \eqref{eq:mesa_updates_x}. This optimization problem is the same as the classic SLAM NLS optimization problem \eqref{eq:cslam_as_nls} with additional terms: 

\begin{equation}
    \label{eq:biased_prior}
    \frac{\beta}{2}\norm{\logmap{\theta \ominus z} + \frac{\lambda}{\beta}}^2   
\end{equation}

These terms are referred to as "biased priors"~\cite{choudhary_admm_2015}. Biased priors are the mechanism by which MESA constrains all robots' local solutions to be equal for any shared variables -- thus ensuring a consistent state estimate for the team. Importantly, biased priors can be represented in standard "factor" from $||h(\Theta) - m||^2_{\Sigma}$ meaning that \eqref{eq:mesa_updates_x} can be solved with existing SLAM optimization methods.

Iterates \eqref{eq:mesa_updates_z}, \eqref{eq:mesa_updates_dual}, and \eqref{eq:mesa_updates_beta} are computationally inexpensive and occur when a pair of robots communicate. The pair first communicates their local solution to any shared variables. Then, each robot updates their edge variables as the interpolation of the local estimates. Edge variables are the values to which biased priors will pull a robot's local solution. Importantly, a pair of robots' edge variables for a shared variable are always equal ($z_{(i,j)_s} = z_{(j,i)_s}$). Therefore, each robot's biased priors will pull their solutions towards a consistent estimate, even as their local solutions change between communications. Next, the pair updates dual variables. The dual variables are an adaptive value by which MESA tightens the equality constraints for variables shared between robots. Finally, robots can optionally update $\beta$ to change the step size for dual variable gradient ascent and further tighten constraints.

It has been shown that via this process, MESA converges to consistent solutions on generic batch C-SLAM problems \cite{mesa_mcgann_2023}. Moreover, MESA is able to converge faster and more accurately than other prior works while permitting asynchronous communication between robots. 

\begin{remark}[MESA Convergence] \label{remark:mesa_convergence}
Prior work has demonstrated empirically that MESA converges for typical SLAM problems~\cite{mesa_mcgann_2023}. However, there is no formal poof of its convergence. Variants of C-ADMM has been proven to converge for non-convex problems~\cite{admm_nonconvex_converge_hong_2016, seminal_admm_nonconvex_converge_wang_2019}, on-manifold problems~\cite{riemannian_admm_li_2023}, problems with nonlinear constraints~\cite{nonlinear_constraints_converge_sun_2023}, asynchronous problems~\cite{o1k_convergence_wei_2013}, and separable problems~\cite{sova_shorinwa_2020}. However, to the best of the author's knowledge, it has not been shown to converge for problems like C-SLAM that exhibit all of these traits and have coupled nonlinear constraints.
\end{remark}
\section{Incremental MESA}\label{sec:methodolody}

In this section we introduce iMESA, an incrementalization of the MESA algorithm developed to solve the the incremental, communication-limited problem outlined in Sec.~\ref{sec:problem_definition}. As discussed in Sec.~\ref{sec:related-work}, batch C-SLAM back-end algorithms cannot be naively applied to the incremental problem. MESA, however, provides an algorithmic structure that permits us to design an effective incrementalized version.

\subsection{iMESA Theoretical Overview}

An iteration of MESA can be broadly broken down into two steps -- a local optimization step \eqref{eq:mesa_updates_x} and a communication step \eqref{eq:mesa_updates_z}, \eqref{eq:mesa_updates_dual}, and \eqref{eq:mesa_updates_beta}. In the local optimization step, robots compute their local estimate, pushing this solution to equal that of the other robots via biased priors. In the communication step, robots share relevant portions of their estimates and tighten shared variable constraints to provide better consistency. It is over many iterations that shared variable constraints are fully realized and a final consistent solution is reached.

The key idea of iMESA is to amortize this process (i.e. the tightening of constraints) over time as communication is available between robots. This amortization is possible with MESA for three reasons. Firstly, and most importantly, intermediate results provided by MESA (i.e. from iterations in which shared variable constraints are not tight) are still practically useful. The solution to \eqref{eq:mesa_updates_x} is effectively lower bounded by that found by a robot operating independently of the team and improved with even partially tight constraints. Secondly, MESA demonstrates fast convergence and therefore consistency and solution quality improve quickly even when the algorithm is amortized over time. Finally, MESA is fully agnostic to the optimization process used to solve \eqref{eq:mesa_updates_x} as it only requires that we have a solution when communication occurs. This permits us to use existing incremental SLAM optimization methods to achieve online performance.

\begin{remark}[Batch C-SLAM Amortization] \label{remark:batch_cslam_amortization}
    This structure of amortization is not compatible with all batch C-SLAM algorithms. Algorithms based on distributed gradient descent~\cite{dgs_choudhary_2017, dc2pgo_tian_2021, asapp_tian_2020, geod_cristofalo_2020} do not provide useful results at intermediate steps of the algorithm. MESA, therefore, provides a unique basis for an incremental C-SLAM back-end.
\end{remark}

From this key idea, we develop a two stage algorithm consisting of a local update step in which a robot incrementally updates \eqref{eq:mesa_updates_x} as new measurements are observed and a communication step in which a pair of robots update their edge and dual variables to tighten shared variable constraints. These steps, combined with some practical bookkeeping, construct the iMESA algorithm and are described in detail below.

\subsection{iMESA Implementation Details}
To realize the idea proposed above, we must handle the practicalities of implementing the algorithm. Firstly, let each robot $i$ maintain the following internal state:

\begin{center}
\renewcommand*{\arraystretch}{1.2}
\begin{tabular}{ l m{0.33\textwidth} }
    \texttt{isam2} & An instance of an iSAM2 optimizer~\cite{isam2_kaess_2012}. \\
    \hline
    $\Theta$ & The local solution. \\
    \hline
    $S_j ~\forall~j\in\Rc$ & Sets variables shared with other robots. \\ 
    \hline
    $S'_j~\forall~j\in\Rc$ & Like $S_j$ but variables are known only to robot $i$ due to limited communication.  \\  
    \hline
    $\Lambda = \{ \lambda_{(i,j)_s}\}$ & The set of all dual variables.   \\
    \hline
    $Z = \{z_{(i,j)_s}\}$ & The set of all edge variables.\\
    \hline
    $B = \{\beta_{(i,j)_s}\}$ & The set of all penalty parameters.\\
    \hline
    $\Cc$ & Cache of biased priors to be added the optimizer. \\
    \hline
    $\Kc$ & Cache of variables affected by recent communication. \\ 
\end{tabular}
\end{center}

Secondly, let each robot $i$ be configured with the following hyper-parameter values:

\begin{center}
\renewcommand*{\arraystretch}{1.2}
\begin{tabular}{ l c m{0.33\textwidth} }
    $\beta_{uninit}$ & $1e^{-4}$ & Penalty parameter for uninitialized biased priors. \\
    \hline
    $\beta_{init}$ & 1.0 & Penalty parameter for biased priors.
\end{tabular}
\end{center}

The purpose of these state variables and parameters are explored in detail in the following sub-sections.

\subsubsection{Bookkeeping} Unlike batch settings, in incremental C-SLAM, the underlying problem is changing at every timestep as robots take new measurements. We must therefore, efficiently track information required to perform local updates and communicate information between robots.

To facilitate communication, each robot $i$ tracks two sets for all other robots: $S_j~\forall~j\in\Rc$ which holds the set of variables shared between the pair $(i,j)$ and $S'_j \subset S_j$ which holds the set of variables shared between the pair but are currently known only to robot $i$. Such variables exist as robot $i$ may have detected a loop-closure to a variable in robot $j$'s factor-graph in the period of time since their last communication. When a robot $i$ makes a new inter-robot measurement that observes a shared variable not already marked in $\Sc_j$ for the respective robot $j$ we must add it to both $\Sc_j$ and $\Sc'_j$.

For each variable $s$ shared with another robot $j$ each robot $i$ must also maintain a corresponding dual variable $\lambda_{(i,j)_s}$, edge variable $z_{(i,j)_s}$, penalty parameter $\beta_{(i,j)_s}$, and biased prior \eqref{eq:biased_prior}. Thus, when observing a new shared variable a robot must extend $\Lambda$, $Z$, and $B$ as well as construct a corresponding biased prior that references these new values. The bookkeeping process is summarized in Alg.~\ref{alg:imesa_bookkeep}.

\begin{algorithm}
\footnotesize 
\captionsetup{font=footnotesize} 
\caption{iMESA: Bookkeep (Local to robot $i$)}\label{alg:imesa_bookkeep}
    \begin{algorithmic}[1]
    \State \textbf{In:} New shared variables $\Sc_{new}$ and their initial estimates $\Phi$
    \For{each variable $s$ in $\Sc_{new}$ shared with robot $j$}
        \State Extend $\Sc_j$ and $\Sc'_j$ for new variable $s$
        \State Extend $\Lambda$ for $j,s$ with value $\mathbf{0}$
        \State Extend $Z$ for $j,s$ with value $\Phi_s$
        \State Extend $B$ for $j,s$ with value $\beta_{uninit}$
        \State Extend $\Cc$ with a biased prior on $s$ referencing $\lambda_{(i,j)_s}$, $z_{(i,j)_s}$, and $\beta_{(i,j)_s}$
    \EndFor
    \end{algorithmic}
\end{algorithm}

\begin{remark}[Bookkeeping Initial Values]\label{remark:bookkeeping_init_vals}
All dual variables are constructed with zero vector of appropriate dimension to match the MESA algorithm~\cite{mesa_mcgann_2023}. All edge variables are constructed with the local initial estimate as there has not yet been opportunity to communicate with the robot $j$ to interpolate estimates. Additionally, since we know that $z_{(i,j)_s}$ is constructed to an incorrect value, we construct the penalty parameter with a very small value ($\beta_{uninit}$) so that the biased prior does not yet have an impact on the robot's local solution.
\end{remark}

\subsubsection{Update}
When new factors are observed, we must not only bookkeep the new information, but also update our local solution. The process of doing so is simple in the context of iMESA as prior work, namely that of Kaess et al., has already developed algorithms for efficient incremental updates to a SLAM problem using the Bayes Tree~\cite{isam2_kaess_2012}. Encapsulating the iSAM2 algorithm, iMESA's update is summarized in Alg.~\ref{alg:imesa_update}

\begin{algorithm}
\footnotesize 
\captionsetup{font=footnotesize} 
\caption{iMESA: Update (Local to robot $i$)}\label{alg:imesa_update}
    \begin{algorithmic}[1]
    \State \textbf{In:} New factors $\Fc$ and new initial estimates $\Phi$
    \State $\Sc_{new} \gets$ any new shared variable observed in $\Fc$
    \State Bookkeep($\Sc_{new}, \Phi$) \Comment{Alg.~\ref{alg:imesa_bookkeep}}
    \State $\Theta \gets$  \texttt{isam2.update}$(\Fc \cup \Cc, \Phi,$ \texttt{reelim=}$\Kc)$ \Comment{\cite[Alg.~8]{isam2_kaess_2012}}
    \State Reset $\Cc \gets \varnothing$ and $\Kc\gets\varnothing$
    \end{algorithmic}
\end{algorithm}

\begin{remark}[Biased Prior Cache]
    During Alg.~\ref{alg:imesa_update}, we incorporate any biased priors in $\Cc$ that may have been cached during communication(s) since the last update. Biased priors are cached and included in the update here to avoid repeated work that would be required by updating iSAM2 explicitly after communication. This delays effects of communication from appearing in the local solution, but the delay is minor given we expect updates to be frequent. If effects from communication are required immediately, the iSAM2 update can be called both here and below in Alg.~\ref{alg:imesa_communication} at a computational cost.
\end{remark}

\begin{remark}[iSAM2 Implementation Details]
    By default iSAM2 will only re-linearize factors for variables which have changed by more than a set threshold. However, since edge and dual variables are not considered variables by iSAM2 (as they are not optimized by iSAM2), we must specifically indicate that biased priors who's edge and dual variables have been updated must be re-eliminated. We use the iSAM2 implementation provided in \texttt{gtsam}~\cite{dellaert_gtsam_tech_report_2012} that provides this functionality during iSAM2's update. Specifically, we mark for re-eliminated all variables $\Kc$ for which corresponding dual and edge variables have been modified since the last update.
\end{remark}

\subsubsection{Communication} When communication is available between two robots, MESA only requires that the pair share their current solution for all shared variables. This requires minimal bandwidth on the communication link between the pair. However, as noted above, the set of shared variables may have changed since the pair's last communication and each robot may be only partially aware of these changes.

Therefore, iMESA makes use of a two stage communication process. First, the robots must update each-other on their set of shared variables by communicating $S'_j$. Upon receiving $S'_i$ from the other robot, each must then update their bookkeeping. After this, both robots can share their local estimates for the, now jointly known, set of shared variables and update $\Lambda$, $Z$, and $B$. This communication process is summarized in Alg.~\ref{alg:imesa_communication}.

\begin{algorithm}
\footnotesize 
\captionsetup{font=footnotesize} 
\caption{iMESA: Communication (Robot $i$ communicating with robot $j$)}\label{alg:imesa_communication}
    \begin{algorithmic}[1]
    \State Send $\Sc'_j$ to robot $j$ and receive $\Sc'_i$ from robot j \Comment{Comm. Step 1}
    \State Bookkeep($\Sc'_i$, $\Theta$) and reset $\Sc'_j = \varnothing$ \Comment{Alg.~\ref{alg:imesa_bookkeep}}
    \State Send $\theta_{s_i}$ and receive $\theta_{s_j}~\forall~s\in S_j$\Comment{Comm. Step 2}
    \For{$s$ in $\Sc_j$}
        \State Update $z_{(i,j)_s} \in Z$ using Eq.~\ref{eq:mesa_updates_z}
        \State Update $\lambda_{(i,j)_s} \in \Lambda$ using Eq.~\ref{eq:mesa_updates_dual}
        \State \textbf{If} $\beta_{(i,j)_s} == \beta_{uninit}$ \textbf{then} update $B$ using $\beta_{(i,j)_s} \gets \beta_{init}$ 
        \State Extend $\Kc$ with $s$
    \EndFor
    \end{algorithmic}
\end{algorithm}

\begin{remark}[Two Stage Communication]\label{remark:two_stage_comms}
    A two stage communication process, is required for efficient communication by any incremental C-SLAM back-end, including DDF-SAM2 and DLGBP in scenarios with sparse communication. All incremental C-SLAM back-ends communicate some information pertaining to shared variables. Therefore, before communicating, robots must first agree upon a joint set of shared variables which may have changed since the last communication. Alternatively, robots could simply send their entire solution, but doing so would induce a much greater communication cost.
\end{remark}

\subsubsection{Penalty Parameters \& Biased Priors}
A keen reader will notice that the penalty parameter update in Alg.~\ref{alg:imesa_communication} differs significantly from that used in MESA \eqref{eq:mesa_updates_beta}. Updating the penalty parameter along with updating dual variables can be used to tighten the shared variable constraints. However, unlike dual variables, changes to the penalty parameter are not adaptive and once tightened, they cannot be loosened. This can cause challenges in incremental settings when loop-closures are observed. If constraints are enforced with a large penalty parameter, it can prevent a robot's local optimization process from updating its solution with that new information. Dual variables on the other hand, adapt in both magnitude and direction as local estimates change. Therefore, we use only dual variable updates to tighten constraints and use a constant penalty parameter of $\beta = 1$ once biased priors are initialized. Note that this requires we maintain a unique penalty parameter $\beta_{(i,j)_s}$ for each shared variable $s$.

However, while very general, the selection of $\beta=1$ can cause stability issues in some edge cases. With an initial dual variable ($\lambda = \mathbf{0}$), a biased prior is effectively a standard prior factor with an isotropic noise model ($\Sigma = I$). Therefore, biased priors that affect pose variables provide a weak constraint on the rotational component. This can cause numerical stability issues when inter-robot measurements don't observe rotations (e.g. bearing+range). The crux of the challenge is that pose translation ($t$) and rotation ($r$) components live in different domains with different scales $t \in [-D, D]$ and $r\in (-\pi, \pi]$ where $D$ is the size of the team's operational area. To improve stability, we also provide biased priors on poses with a noise model so they have similar effect between the two domains. We specifically use a noise model with $\sigma_r = 0.1$ and $\sigma_t = 1$.

\subsubsection{iMESA Summary}
The iMESA algorithm consists entirely of running Alg.~\ref{alg:imesa_update} and Alg.~\ref{alg:imesa_communication} on demand when measurements and communications are available. These algorithms are very efficient and allow iMESA to run online using sparse, bandwidth-limited communication as outlined in Sec.~\ref{sec:problem_definition}. A visual depiction of iMESA in operation can be found in Fig.~\ref{fig:amortization_explination}.

\begin{remark}[iMESA Convergence]
Given that MESA does not come with any convergence guarantees (See Remark~\ref{remark:mesa_convergence}), iMESA also does not guarantee convergence. This, however, is standard among incremental distributed SLAM algorithms. As noted in Sec.~\ref{sec:related-work}, DDF-SAM2 does not enforce consistency between shared variables and, in-turn, will not converge to a single solution~\cite{ddfsam2_cunningham_2013}. Additionally, being an extension of Loopy Belief Propagation, DLGBP is not guaranteed to converge for even some convex problems, let alone non-convex C-SLAM problems \cite{loopy_bp_murphy_1999}.
\end{remark}

\begin{remark}[Relationship to iDFGO]\label{remark:idfgo_imesa_differences}
Concurrent to this work Matsuka and Chung also proposed an incremental factor-graph optimization algorithm based on C-ADMM that can be applied to C-SLAM problems~\cite{idfgo_matsuka_2023}. This algorithm (iDFGO) and iMESA share a similar spirit -- to use C-ADMM for distributed factor-graph optimization and utilize existing incremental optimizers for efficient local optimization. However, these algorithms differ significantly.

Unlike iMESA, iDFGO assumes that agents perform multiple iterations per-timestep where each iteration requires synchronized communication over a connected network. This design prevents iDFGO from meeting the communication requirements outlined in Sec.~\ref{sec:problem_definition}. Additionally, iDFGO was primarily designed as a convex algorithm, and in-turn formulates constraints linearly. This is equivalent to using Chordal constraints which were found to result in significantly worse performance on C-SLAM problems as compared to Geodesic constraints that are used in iMESA~\cite{mesa_mcgann_2023}. Further, unlike iMESA, iDFGO does not handle online bookkeeping and instead assumes that agents have global knowledge of which variables are shared. Such global knowledge is impossible to obtain in incremental C-SLAM settings. Therefore, unlike iMESA, iDFGO's properties make it difficult to apply to the incremental, communication limited C-SLAM problems that we address in this paper.
\end{remark}
\section{Experiments}\label{sec:experiments}
In this section we evaluate the performance of iMESA on a variety of real and synthetic C-SLAM problems and demonstrate that it is able to achieve superior performance to state-of-the-art incremental, distributed C-SLAM back-ends. 

\subsection{Experimental Setup}

\subsubsection{Prior Works \& Baselines}\label{sec:exp:prior_works_and_baselines}
In all of our experiments we compare the performance of iMESA to the incremental distributed C-SLAM solvers DDF-SAM2~\cite{ddfsam2_cunningham_2013} and DLGBP~\cite{robot_web_journal_murai_2024}. DDF-SAM2 was implemented by the authors of this work and uses the Naive-Bayes approximation as suggested in the original paper. Implementation of DLGBP was provided by the original authors. We evaluate against standard DLGBP as well as Windowed DLGBP with window size $w = 30$.

In addition to these prior works we also validate iMESA against two baselines. The first is a centralized solver that aggregates all measurements into a single system and solves it incrementally using the iSAM2 optimizer~\cite{isam2_kaess_2012}. The second is an independent solver in which all robots use iSAM2 to solve their local factor-graph without any inter-robot collaboration.

\subsubsection{Synthetic Datasets}\label{sec:exp:dataset_gen}
\begin{figure}[t]
    \centering
    \begin{subfigure}{0.32\linewidth}
      \centering
      \includegraphics[width=0.9\linewidth,trim={1cm 1.5cm 2cm 1.5cm},clip]{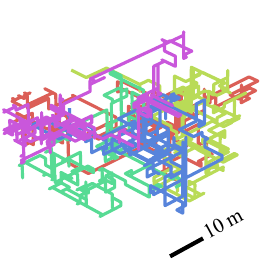}
      \vspace{-0.25cm}
      \caption{}
      \label{fig:example_trajectories:models_3d}
    \end{subfigure}%
    \begin{subfigure}{0.32\linewidth}
      \centering
      \includegraphics[width=0.9\linewidth]{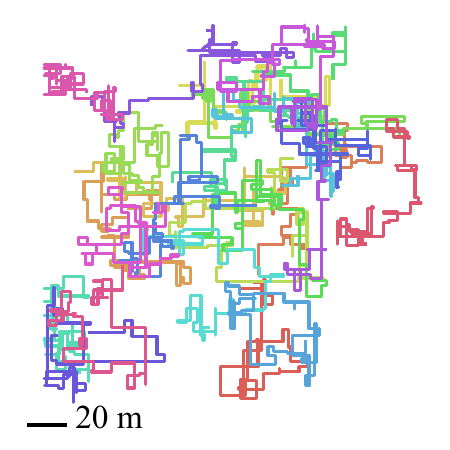}
      \vspace{-0.25cm}
      \caption{}
      \label{fig:example_trajectories:scale_25}
    \end{subfigure}%
    \begin{subfigure}{0.32\linewidth}
      \centering
      \includegraphics[width=0.9\linewidth]{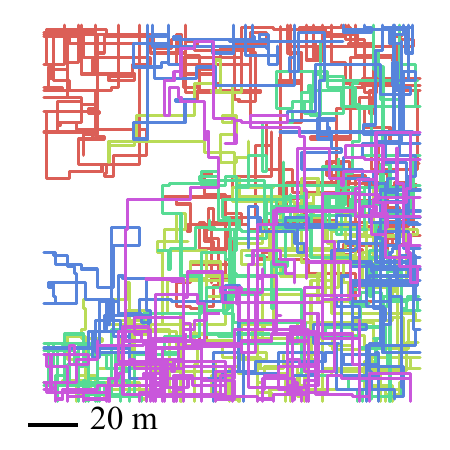}
      \vspace{-0.25cm}
      \caption{}
      \label{fig:example_trajectories:lifelong_5000}
    \end{subfigure}%
    \caption{Example ground-truth synthetic datasets: (a) 3D PGO dataset from Exp.~\ref{sec:exp:models}, (b) 25 robot 2D dataset from Exp.~\ref{sec:exp:scale}, and (c) 5000 length 2D dataset from Exp.~\ref{sec:exp:lifelong}. Each color is the trajectory of a different robot.}
    \label{fig:example_synthetic_datasets}
    \vspace{-0.25cm}
\end{figure}

We evaluate the above methods on synthetic and real-world datasets. Synthetic datasets are used to explore the effect of problem conditions on algorithm performance. Synthetic datasets are generated in both 2D and 3D by randomly sampling odometry from a categorical distribution with options of $1m$ forward motion as well as $\pm 90^\circ$ rotation around each available axis. Intra-robot loop closures are detected whenever a robot returns within $2m$ of a previous pose and added with probability $p=0.4$. Intra-robot loop closures are always modeled as relative pose measurements. Inter-robot loop-closures are added every $5$ timesteps to robots within $30m$ of each other. Inter-robot loop-closures can be modeled as relative pose measurements, bearing-range measurements, or range measurements. Visualizations of example synthetic datasets are provided in Fig.~\ref{fig:example_synthetic_datasets}.

\begin{remark}[Synthetic Data Noise Models]
    All synthetic datasets use diagonal noise models with the following standard deviations. Priors: $(\sigma_r= 0.1^\circ, \sigma_t=0.01m)$, Odometry: $(\sigma_r= 1^\circ, \sigma_t=0.05m)$, Relative Pose Loop-Closures: $(\sigma_r=1^\circ, \sigma_t=0.1m)$, Bearing+Range: $(\sigma_b=1^\circ, \sigma_d=0.1m)$, Range: $(\sigma_d=0.1m)$. Where the standard deviations denote the measurement component to which they apply: $r$ for rotation, $t$ for translation, $b$ for bearing, and $d$ for distance/range. 
\end{remark}

\subsubsection{Communication}\label{sec:exp:comm_model}
For all experiments we simulate communication connectivity between robots. Robots attempt to communicate at each timestep of the algorithm. Robots are only permitted a single pairwise communication per timestep to any robot within a communication range of $30m$. Additionally, we randomly drop $10\%$ of communications to simulate instability in robots establishing a connection.

\subsubsection{Metrics}\label{sec:exp:metrics}
We evaluate performance of all methods using Average Trajectory Error (ATE) to inspect the practical quality of results. However, we look to evaluate this performance not only for the final timestep, but, for intermediate timesteps during which a real robotic system will need accurate state estimates for downstream tasks. To measure this performance, we use an incremental variant of ATE~\cite[Eq.~3]{mcgann_risam_2023}. We compute iATE independently for each robot and sum all errors into two values, summarizing translational and rotational accuracy of the team. In all experiments iATE is reported in meters and radians for translational and rotational error respectively.

\subsection{Measurement Models Experiment} \label{sec:exp:models}
In our first experiment we investigate how iMESA and prior works handle different problem formulations. Specifically we investigate 2D and 3D PGO, as well as 2D C-SLAM with bearing+range and range only inter-robot measurements. These different scenarios have different observability and nonlinearities, making them each an interesting optimization challenge. We run each method on $20$ random datasets with $5$ robots, each traversing a trajectory $1000$ poses long. 

\begin{figure}[t]
    \centering
    \includegraphics[width=1.0\linewidth]{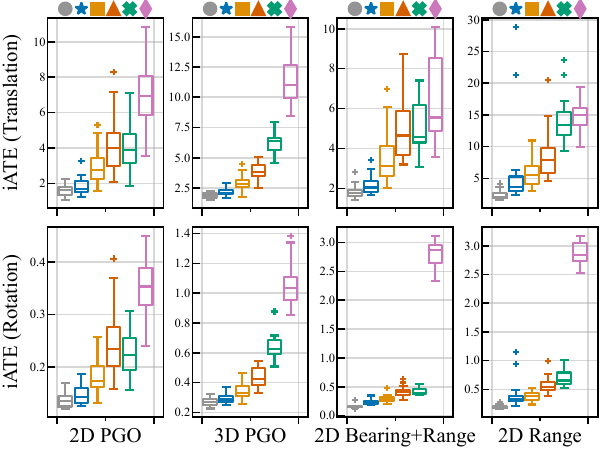}
    \caption{ 
        iATE for our proposed algorithm iMESA~(\SymiMESA) along with centralized~(\SymCentralized) and independent~(\SymIndependent) baselines and prior works DDFSAM2~(\SymDDFSAM), DLGBP~(\SymDLGBP), and Windowed DLGBP~(\SymDLGBPWindowed) on different C-SLAM formulations. Across problem formulations iMESA provides accurate results and outperforms state-of-the-art prior works.
    }
    \label{fig:models_experiment_results}
    \vspace{-0.25cm}
\end{figure}

We summarize the iATE achieved on each trial via a box and whisker plot in Fig.~\ref{fig:models_experiment_results}. There, we see that iMESA is able to outperform all prior works across problem scenarios and recover the closest solution to that of a centralized optimizer. Further, we can see Windowed DLGBP performs significantly worse than its non-windowed variant due to the presence of loop-closures in the datasets. Finally, it is notable that in the range scenario, iMESA, Windowed DLGBP, and DDF-SAM2 reported outlier performance on a small number of trials, likely due to the significant nonlinearity of range measurements.

\subsection{Problem Scale Experiment}\label{sec:exp:scale}
Next we evaluate how the methods perform as the size of the team grows. For each team size we generate $20$ random 2D PGO datasets with robots traversing trajectories of $500$ poses. 

\begin{figure}[t]
    \centering
    \includegraphics[width=1.0\linewidth]{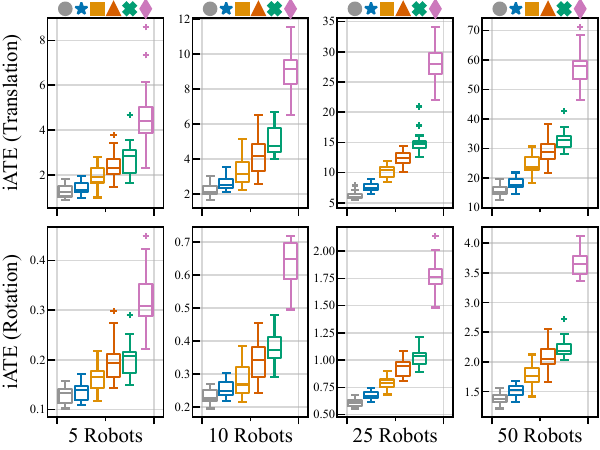}
    \caption{
        iATE for our proposed method iMESA~(\SymiMESA) along with centralized~(\SymCentralized) and independent~(\SymIndependent) baselines and prior works DDFSAM2~(\SymDDFSAM), DLGBP~(\SymDLGBP), and Windowed DLGBP~(\SymDLGBPWindowed) as robot team size increases. Our proposed algorithm iMESA outperforms prior works and all methods provide consistent performance as team size grows.
    }
    \label{fig:scale_experiment_results}
    \vspace{-0.25cm}
\end{figure}

\begin{figure}[t]
    \centering
    \includegraphics[width=1.0\linewidth]{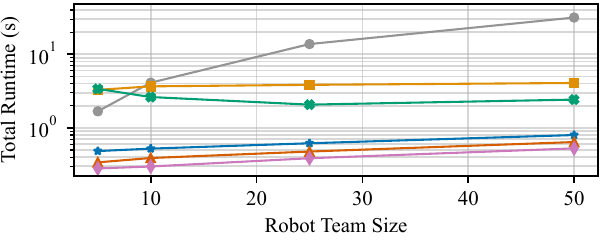}
    \caption{ 
        Total runtime for our proposed method iMESA~(\SymiMESA) along with centralized~(\SymCentralized) and independent~(\SymIndependent) baselines and prior works DDFSAM2~(\SymDDFSAM), DLGBP~(\SymDLGBP), and Windowed DLGBP~(\SymDLGBPWindowed) as team size increases. All distributed methods scale well as team size increases, however, the centralized solver sees significantly increased runtime as team size grows. 
    }
    \label{fig:scale_runtime_results}
    \vspace{-0.25cm}
\end{figure}

We summarize results from these trials in Fig.~\ref{fig:scale_experiment_results} in which we can observe that again, iMESA outperforms prior works. We can also see that the relative performance of all distributed methods is consistent across scales. We note that the error increases as team size increases as we sum errors from all robots (See.~\ref{sec:exp:metrics}). We can gain additional insight by looking at the trend in runtime across team size. In Fig.~\ref{fig:scale_runtime_results}, we plot the the total runtime averaged over all trials. Here, we can see that unlike the distributed methods, the centralized baseline computational requirements scale very poorly with team size. These results add evidence to the claim that centralized approaches are difficult to employ with large multi-robot teams~\cite{subt_lessons_2023}. The distributed methods, on the other hand, all scale very well with the size of the team. 

\subsection{Long-term Operation Experiment}\label{sec:exp:lifelong}
In our next synthetic experiment we investigate how the methods handle long-term operation. For this experiment we evaluate on $20$ random 2D PGO datasets consisting of $5$ robots traversing trajectories of increasing length.

\begin{figure}[t]
    \centering
    \includegraphics[width=1.0\linewidth]{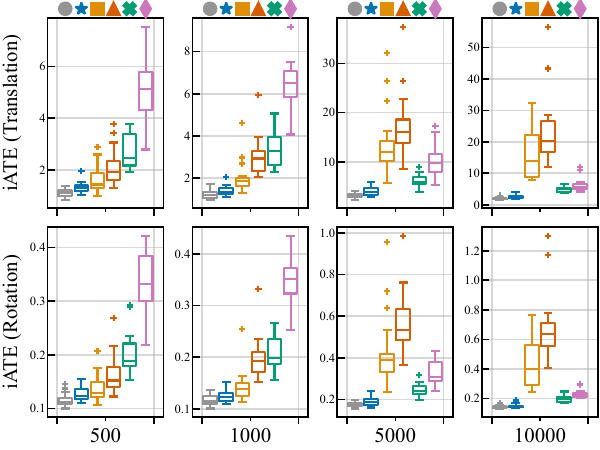}
    \caption{
        iATE for our proposed method iMESA~(\SymiMESA) along with centralized~(\SymCentralized) and independent~(\SymIndependent) baselines and prior works DDFSAM2~(\SymDDFSAM), DLGBP~(\SymDLGBP), and Windowed DLGBP~(\SymDLGBPWindowed) as operational time increases. Our algorithm iMESA outperforms prior works across all lengths. Interestingly, DLGBP and its windowed variant see degraded performance on longer problems.
    }
    \label{fig:lifelong_experiment_results}
    \vspace{-0.25cm}
\end{figure}

Long-term operation performance is summarized in Fig.~\ref{fig:lifelong_experiment_results}. In this figure we can see the continued trend that iMESA outperforms existing distributed C-SLAM back-ends and achieves results closest to a centralized solver. Interestingly, we can see in this experiment that DLGBP and its windowed variant perform worse as the trajectory length grows. We hypothesize that DLGBP struggles to properly incorporate very long term loop-closure information even without windowing. However, the exact cause of this degradation in performance was not studied in depth and deserves focus in future work.

\begin{figure}[t]
    \centering
    \includegraphics[width=1.0\linewidth]{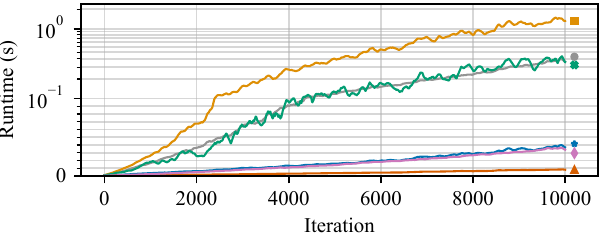}
    \caption{
        Iteration runtime for our proposed method iMESA~(\SymiMESA) along with centralized~(\SymCentralized) and independent~(\SymIndependent) baselines and prior works DDFSAM2~(\SymDDFSAM), DLGBP~(\SymDLGBP), and Windowed DLGBP~(\SymDLGBPWindowed) as the factor-graph size increases. Our algorithm iMESA achieves very efficient performance requiring only marginally more computation time per iteration than locally running iSAM2 (independent). DLGBP iteration times grow significantly indicating that it may be difficult to apply to real-world applications. Note: the y-axis scale is linear up to $0.1$ second and log scale above that threshold.
    }
    \label{fig:lifelong_runtime_results}
    \vspace{-0.25cm}
\end{figure}

We also use the results from this experiment to evaluate the feasibility of online performance for all methods. Using results from the $10k$ length trials, we compute the iteration run-time averaged across any parallel computation (i.e. across robots for all methods except the centralized approach). Iteration runtime includes all computation required by updates and during any communication performed at that iteration. However, it does not include network transmission time that would be required during actual communication. We further average these run-times across all $20$ trials and plot the results in Fig.~\ref{fig:lifelong_runtime_results}.

These iteration runtime results demonstrate that iMESA is very efficient. Even with $10k$ poses, the average iteration time remains less than $0.04$ seconds requiring little more time than standard iSAM2 on problems of the same size (i.e. independent approach). This indicates that iMESA will be able to achieve real-time performance even in long-duration applications. To contextualize this result, we also run MESA on a single $10k$ length dataset. To optimize only the final iteration, MESA required over 800 pairwise communications and over $2$ minutes of computation on each robot. This highlights the importance of incrementalization and the difficulties one would face if naively applying batch algorithms to incremental problems. These results also show that, as expected, Windowed DLGBP is the fastest. We note that it's iteration runtime does increase despite being theoretically constant as larger data structures require more time to update. Further we can see that DLGBP is by far the most expensive, with final iteration times well over $1$ second. While DLGBP may be feasible for small incremental C-SLAM problems, it will likely not achieve real-time performance for larger problems.

\subsection{Delayed Communication Experiment}
In all experiments we simulate asynchronous communication between robots by limiting communication range and restricting robots to pairwise communication which causes robots to always be working with some outdated information. One aspect of communication networks that this does not explore is delays that may occur when communicating over networking infrastructure. In this experiment we explore the effect of delays on the performance of iMESA. We induce gaps in communication by restricting robots to communication only every other timestep and increase communication failure rate to $0.4$. Further, robots communicate their oldest complete information up to a maximum delay of $k$ timesteps. This experiment is run on $20$ random PGO datasets consisting of $8$ robots traversing trajectories 500 poses long.

\begin{figure}[!t]
    \centering
    \includegraphics[width=0.85\linewidth]{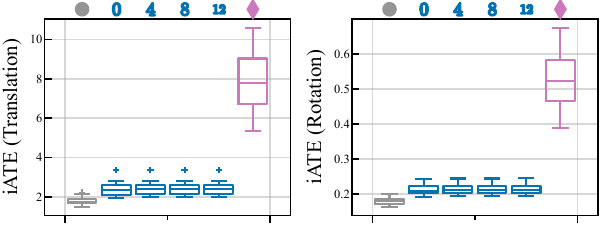}
    \caption{
        iATE for our proposed method iMESA with different amounts of maximum communication delay [\textcolor{sns:blue}{\textbf{0}}, \textcolor{sns:blue}{\textbf{4}}, \textcolor{sns:blue}{\textbf{8}}, \textcolor{sns:blue}{\textbf{12}}] along with centralized~(\SymCentralized) and independent~(\SymIndependent) baselines. Even with increasing delays to some information, iMESA is able to perform consistently and provide accurate solutions. 
    }
    \label{fig:delay_experiment_results}
    \vspace{-0.2cm}
\end{figure}

We summarize the results that iMESA achieves under increasing maximum delay and summarize the iATE achieved in Fig~\ref{fig:delay_experiment_results}. In this figure, we can see that iMESA maintains consistent performance across scales of maximum delay. This indicates that iMESA will be able to tolerate the networking delays we may encounter in real-world deployments. We also note that iMESA performance gap to the centralized solution is slightly larger than in other experiments due to the slower communication rate and increased communication dropout.

\subsection{Real-Data Multi-Robot Experiment}

In our final experiment we explore how incremental distributed C-SLAM methods perform on real-world data. We use the open-source Nebula Autonomy multi-robot datasets. 
These datasets consist of LiDAR data and 3D pose-graphs generated by LAMP 2.0 in subterranean tunnel environments~\cite{lamp2_chang_2022}. We replay these pose-graphs incrementally for the algorithms investigated in this work, simulating communication using the method described in Sec.~\ref{sec:exp:comm_model} with a larger communication range of $100m$. Additionally, due to the real-world nature of these datasets some loop-closures in the pose-graphs are outliers. In this work we do not focus on handling outliers and instead preprocess datasets to remove outlier measurements.

\begin{figure}[t]
    \centering
    \begin{subfigure}{0.5\linewidth}
      \centering
      \includegraphics[width=1\linewidth,trim={0cm 1cm 0cm 1cm},clip]{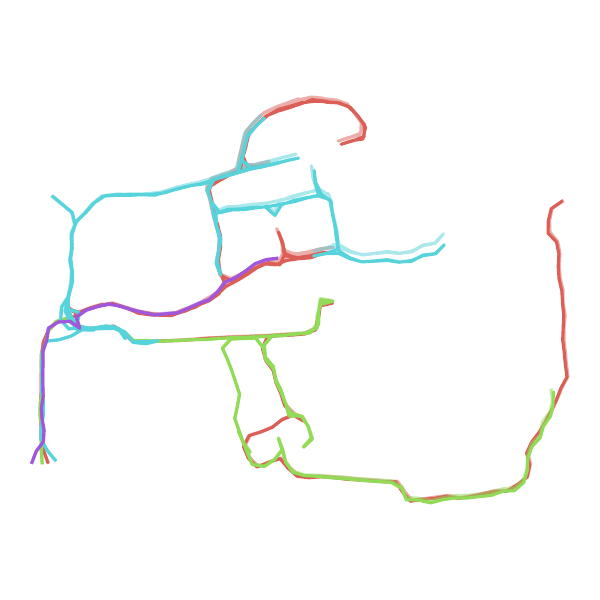}
      \label{fig:nebula_results:traj}
    \end{subfigure}%
    \begin{subfigure}{0.5\linewidth}
      \centering
      \includegraphics[width=1\linewidth,trim={0cm 1cm 0cm 1cm},clip]{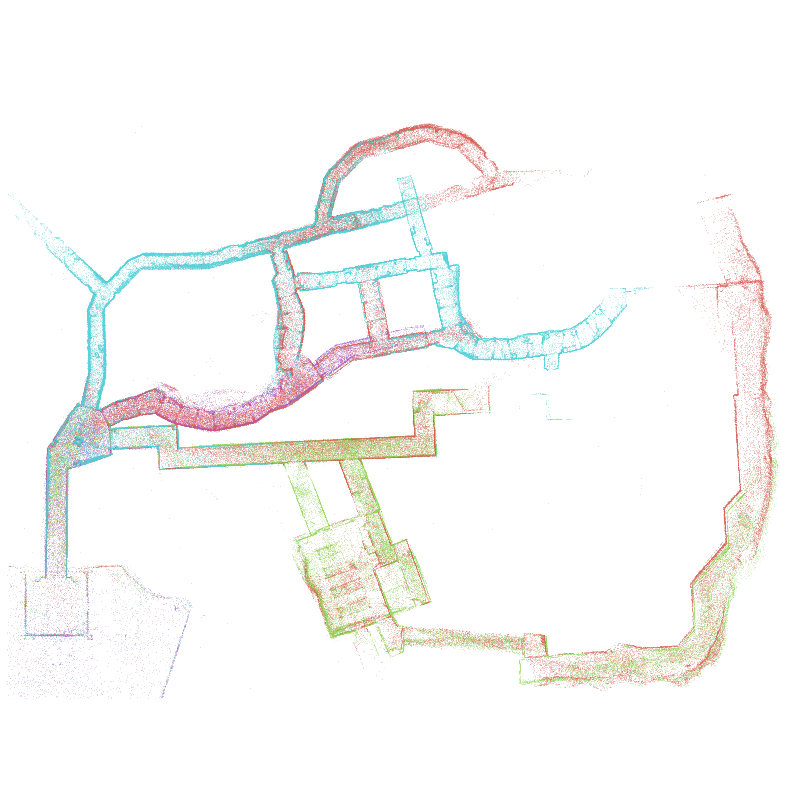}
      \label{fig:nebula_results:map}
    \end{subfigure}%
    \vspace{-0.2cm}
    \caption{Trajectory (left) and corresponding map (right) computed by iMESA on the \texttt{finals} dataset. Different colors denote different robots and lighter colors on trajectory plot depict groundtruth.}
    \label{fig:example_nebula_results}
    \vspace{-0.25cm}
\end{figure}

Accuracy results for all methods on the four datasets are summarized in Table~\ref{tab:nebula_results}. In this table we can see that iMESA performs consistently with results comparable to the centralized solver. We can also see this qualitatively in iMESA's solution to the \texttt{finals} dataset in Fig.~\ref{fig:example_nebula_results}. Interestingly, we found that DLGBP diverged on the \texttt{urban} and \texttt{finals} dataset. This was our only observation of this behavior, but indicates DLGBP is sensitive to problem conditions. 

\begin{table}[t] 
\centering
\renewcommand*{\arraystretch}{1.2}
\caption{iATE (Translation, Rotation) on the Nebula Multi-Robot Datasets}
\vspace{-0.2cm}
\label{tab:nebula_results}
\begin{tabular}{m{0.17\linewidth}c@{\hspace{5pt}}c|c@{\hspace{5pt}}c|c@{\hspace{5pt}}c|c@{\hspace{5pt}}c}
\multirow{2}{*}{Algorithm} & \multicolumn{8}{c}{Dataset}\\
\cline{2-9}
&\multicolumn{2}{c|}{\texttt{tunnel}} & \multicolumn{2}{c|}{\texttt{urban}} & \multicolumn{2}{c|}{\texttt{finals}} & \multicolumn{2}{c}{\texttt{KU}}\\
\hline
Centralized     & 1.48 & 0.093 & 1.20 & 0.076 & 0.53 & 0.096 & 3.05 & 0.118 \\
\hline
iMESA           & \textbf{1.46} & 0.091 & \textbf{1.20} & \textbf{0.071} & \textbf{0.65} & \textbf{0.094} & 3.14 & \textbf{0.117} \\
DLGBP           & 1.91 & 0.093 & --   & --    & --   & --    & 3.37 & 0.120 \\
WDLGBP  & 1.71 & \textbf{0.086} & 1.23 & 0.075 & 0.85 & 0.142 & 4.85 & 0.140 \\
DDFSAM2         & 1.59 & 0.095 & 1.63 & 0.094 & \textbf{0.65} & \textbf{0.094} & \textbf{3.00} & \textbf{0.117} \\
Independent     & 1.64 & 0.102 & 1.38 & 0.083 & 0.71 & 0.099 & 3.71 & 0.124 \\
\end{tabular}
\vspace{-0.2cm}
\end{table}

Finally, we want to note that some distributed methods appear to outperform the centralized solver in this experiment. This occurs as measurements are derived from real-data, and therefore the optimum of \eqref{eq:cslam_as_nls} does not necessarily correspond to the groundtruth trajectory. As such, slight deviations from this optimum that distributed methods exhibit can marginally improve the trajectory error with respect to the groundtruth.

\section{Conclusion and Future Work}\label{sec:conclusion}

In this paper we present iMESA, a novel incremental distributed C-SLAM algorithm. iMESA is able to achieve real-time performance and accurate results with only sparse communication between robots. We demonstrate that iMESA: can handle generic C-SLAM problem formulations, enabling it to be used across application domains, can scale to large robot teams making it useful as future applications increase team sizes, can handle long-term operations and therefore will be applicable to field robotic applications, and outperforms existing incremental distributed C-SLAM back-ends. 

However, iMESA requires additional work to see success in the real-world. iMESA, like other NLS based C-SLAM solvers, is sensitive to outlier measurements. Future work should investigate methods to make iMESA robust to both local and inter-robot outlier measurements. Additionally, to enforce invariants, iMESA requires that the two-stage communication process (Alg.~\ref{alg:imesa_communication}) be fully completed if started. Future work should look to add algorithmic robustness for the case that communication drops out between robots while this process is occurring. Finally, while iMESA has empirically demonstrated quality performance across C-SLAM problem scenarios, future work should look to provide theoretical guarantees on its convergence properties and investigate iMESA's ability to run on-board multi-robot teams operating in the field.

In addition to future work on the iMESA algorithm, we will also note the need for additional investigation into DLGBP based methods. These methods remain promising, but this work found multiple instances where they struggled to converge or diverged entirely. Future work to explain and resolve this behavior would greatly benefit the C-SLAM community.

\section{Acknowledgements}
This work was partially supported by NASA award 80NSSC22PA952 and the NSF Graduate Research Fellowship Program.
The authors would also like to thank Riku Murai for providing their implementation of DLGBP and their insightful discussion regarding Loopy Belief Propagation.

\bibliographystyle{ieeetr}
\footnotesize

\bibliography{references}

\end{document}